\documentclass[10pt, a4paper]{article}

\usepackage{comment}
\usepackage{booktabs}

\usepackage[ruled,vlined]{algorithm2e}
\usepackage{algorithmic}

\usepackage{cuted}
\usepackage{listings}

\usepackage{paracol}

\usepackage[final]{lrec2026} 

\title{Automated Analysis of Global AI Safety Initiatives: A Taxonomy-Driven LLM Approach}

\name{Takayuki Semitsu, Naoto Kiribuchi, Kengo Zenitani} 

\address{Japan AI Safety Institute \\
         Tokyo, Japan \\
         }

\abstract{
We present an automated crosswalk framework that compares an AI safety policy document pair under a shared taxonomy of activities. Using the activity categories defined in Activity Map on AI Safety as fixed aspects, the system extracts and maps relevant activities, then produces for each aspect a short summary for each document, a brief comparison, and a similarity score. We assess the stability and validity of LLM-based crosswalk analysis across public policy documents. Using five large language models, we perform crosswalks on ten publicly available documents and visualize mean similarity scores with a heatmap. The results show that model choice substantially affects the crosswalk outcomes, and that some document pairs yield high disagreements across models. A human evaluation by three experts on two document pairs shows high inter-annotator agreement, while model scores still differ from human judgments. These findings support comparative inspection of policy documents.
 }

\begin{document}

\maketitleabstract

\section{Introduction}
\label{sec:introduction}

\begin{figure*}[t]
  \centering
  \includegraphics[width=0.95\linewidth]{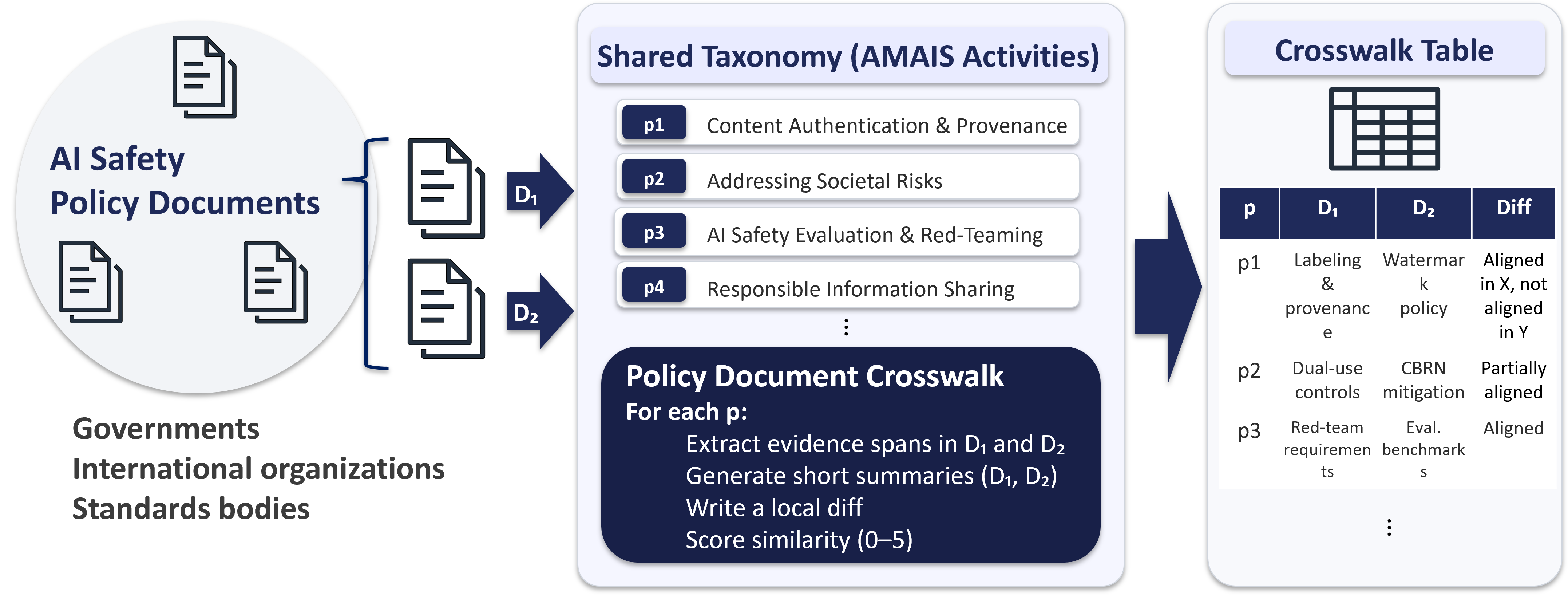}
  \caption{Overview of the crosswalk for policy documents. AI safety policy documents issued by different institutions are compared using a shared taxonomy. For each aspect, the output includes a summary of each document, a comparison, and a similarity score.}
  \label{fig:crosswalk-overview}
\end{figure*}

AI safety policy documents have been produced by governments, international organizations, and standardization bodies, spanning high-level principles and statements as well as more concrete regulations, guidelines, and technical reports. As the variety of document types, granularity, and coverage increases, many documents do not share a unified structure or vocabulary. In this setting, comparative analysis across heterogeneous documents becomes increasingly important for understanding alignments and differences.

A practical comparative method is a \emph{crosswalk}: organizing two documents under a common set of aspects to make similarities and differences visible. However, existing crosswalk practices often rely on ad hoc aspect sets tailored to a specific purpose, without an explicit shared taxonomy that can be reused as a coordinate system for interoperability-oriented comparisons. Moreover, interoperability is ideally extensible beyond two stakeholders, which motivates establishing a robust method even for comparisons between two documents as a foundation.

This paper proposes a framework for automated crosswalk analysis between two AI safety policy documents. We explicitly position a shared taxonomy as a suite of aspects for conducting grounded crosswalk analysis. For the aspect set (p1, \dots, p15) used in this paper, we adopt the \emph{Activities on AI Safety} defined in the \emph{Activity Map on AI Safety} (AMAIS) \citep{amais_aisi_activity_map_ai_safety}. AMAIS is a map that organizes activities on AI safety based on internationally agreed AI-principle documents, namely the Hiroshima AI Process \citep{mic_hiroshima_ai_process} and the Seoul Declaration \citep{seoul_declaration}.

The goal of this study is not to evaluate, rank, or recommend particular policies. Instead, we aim to provide a comparative analysis framework for generating and inspecting crosswalks for each aspect across heterogeneous documents in a standardized and reproducible manner.

The paper makes three contributions:
\begin{enumerate}
  \item \textbf{Demonstrated use of a shared taxonomy for crosswalks:} We use AMAIS activity categories as a reusable coordinate system for organizing comparisons across documents to support interoperability.
  \item \textbf{LLM-based crosswalk analysis:} We define a crosswalk analysis task using activity items that extract and summarize each document under a shared aspect and produces an explicit comparisons.
  \item \textbf{Case study and stability/validity examination:} We apply the framework to a case study of AI safety policy documents and examine stability across models and agreement with human judgments.
\end{enumerate}

\section{Related Work}
\label{sec:related}
Prior work on AI governance has produced comprehensive comparisons across documents of ethics guidelines and national initiatives, typically via manual collection and qualitative coding. Studies have mapped convergences and gaps across large sets of ethics guidelines \citep{Jobin2019}, and systematically compared heterogeneous policy instruments along multiple dimensions \citep{Batool2025}. Other reviews characterize the landscape of governance initiatives within a jurisdiction \citep{AttardFrost2024}, or quantify discrepancies in how trustworthy AI terminology is used across policy and research communities \citep{ToneyWails2024}. While these analyses clarify what themes recur across documents, the underlying crosswalks are often tailored to a study-specific coding scheme, the granularity and definitions of comparison axes (e.g., principle sets, item lists, keyword sets) vary across studies, making it difficult to connect findings and reuse them within a shared coordinate system for interoperability-oriented comparisons. To facilitate interoperability, our study explicitly fixes a common taxonomy as the basis for crosswalks and presents it as a reusable set of comparison axes across documents.

In NLP, automated analysis of documents has advanced through structured classification and summarization. Polisis demonstrates large-scale labeling and user-facing querying of privacy policies \citep{Harkous2018}. For regulations, EUR-LexSum supports learning-based summarization of long legal texts \citep{Klaus2022}, and controlled summarization aims to ensure coverage of salient entities in policy documents \citep{Singh2024}. For policy document comparison, an LLM-agent interface has been proposed that uses hierarchical topic maps to align segments across a document pair \citep{TytarenkoHTM}. However, these approaches primarily target single-document understanding and do not directly operationalize two-document, aspect-wise alignment and difference generation.

Our formulation is also connected to contrastive and comparative summarization. Surveys highlight the diversity of contrastive task definitions and persistent evaluation gaps \citep{Strohle2024}. Comparative summarization methods generate contrastive and common summaries for paired inputs \citep{Iso2022}, and STRUM extracts facet-wise contrasts from web documents without predefined facets \citep{Gunel2023}. Yet, these comparative settings rarely focus on AI safety policy documents or on taxonomy-grounded mappings designed for interoperability.

Finally, LLM-based summarization raises concerns about factuality and hallucination, especially in multi-document settings \citep{Wang2023,Belem2025}. Motivated by these reliability risks, our study defines an automated two-document crosswalk grounded in the AMAIS activity taxonomy (p1, \dots, p15) \citep{amais_aisi_activity_map_ai_safety}, producing per-aspect summaries for each document and an explicit diff, and evaluates stability across models and agreement with human judgements. 

\section{Method}
\label{sec:method}
In this section, we describe the proposed crosswalk framework for policy documents.
As shown in Figure~\ref{fig:crosswalk-overview}, our method takes a document pair $j=(D_1, D_2)$ and a predefined set of aspects $P$ in a shared taxonomy as inputs.
For each aspect $p \in P$, the framework generates (i) an aspect-wise summary of extracted activities in $D_1$ and $D_2$ respectively, (ii) a brief aspect-wise comparison (commonalities and differences),
and (iii) an LLM-computed similarity score on a 0--5 scale.

We use the AMAIS \citep{amais_aisi_activity_map_ai_safety} activity categories $P=\{p_1, \dots, p_{15}\}$ as the shared coordinate system for crosswalk analysis. Table~\ref{tab:amais-p1-p15} lists the activity items with their names and brief descriptions (from the provided AMAIS category definitions).

Let a document pair be denoted by $j=(D_1, D_2)$ (e.g., $A$ and $B$), and let $p \in P$ denote an AMAIS activity category as an aspect. Among AMAIS categories, each activity item consists of its title, description, and keywords. They are defined in the shared taxonomy. Algorithm~\ref{alg:crosswalk} describes the procedure of the crosswalk. For each method $m \in \{a,b,c,d,e\}$ (corresponding to different LLMs), the automated crosswalk task generates three text outputs and one score:
\begin{itemize}
  \item \textbf{Document~1 aspect summary/extraction} $S^{(1)}_{m,j,p}$: a brief text summarizing/extracting what Document~1 ($D_1$) states with respect to aspect $p$.
  \item \textbf{Document~2 aspect summary/extraction} $S^{(2)}_{m,j,p}$: a brief text summarizing/extracting what Document~2 ($D_2$) states with respect to the same aspect $p$.
  \item \textbf{Aspect-wise diff summary} $S^{(\Delta)}_{m,j,p}$: a brief text describing the difference between the two documents with respect to $p$ (e.g., agreement, only-one-sided mention, differences in strength or means, or contradictions).
  \item \textbf{Diff similarity score} $s^{(\Delta)}_{m,j,p}$: a six-level score on a 0--5 scale, computed by the LLM. 
\end{itemize}


\begin{algorithm}[t]
\caption{Crosswalk procedure}
\label{alg:crosswalk}
\KwIn{Document pair $j = (D_1, D_2)$; aspect set $P$; method $m$}
\KwOut{For each $p \in P$: $S^{(1)}_{m,j,p}$, $S^{(2)}_{m,j,p}$, $S^{(\Delta)}_{m,j,p}$, and $s^{(\Delta)}_{m,j,p}$}
\ForEach{$D \in \{D_1, D_2\}$}{
  Extract activity items from $D$\;
  Map the extracted items to aspects in $P$\;
}
\ForEach{$p \in P$}{
  Generate an aspect-wise summary $S^{(1)}_{m,j,p}$ for $D_1$ under $p$\;
  Generate an aspect-wise summary $S^{(2)}_{m,j,p}$ for $D_2$ under $p$\;
  Compare the two summaries and generate an aspect-wise diff summary $S^{(\Delta)}_{m,j,p}$ w.r.t.\ $p$\;
  Assign a similarity score $s^{(\Delta)}_{m,j,p} \in \{0,\dots,5\}$\;
}
\end{algorithm}

We define the similarity score $s^{(\Delta)}_{m,j,p} \in \{0,\dots,5\}$ as follows (5: nearly identical, 4: largely aligned, 3: partial overlap,
2: limited overlap, 1: slight overlap, 0: mostly different). If at least one document has no activity item mapped to $p$,
we set $s^{(\Delta)}_{m,j,p}=0$.

\begin{table*}[t]
\centering

\scriptsize
\begin{tabular}{p{0.05\linewidth} p{0.33\linewidth} p{0.48\linewidth}}
\hline
\textbf{p} & \textbf{Activity} & \textbf{One-sentence description} \\
\hline
p1  & Content Authentication and Provenance & Develop and deploy trustworthy authentication/provenance mechanisms (e.g., watermarking) so users can identify AI-generated content when technically possible. \\
p2  & Addressing Societal Risks & Address risks in high-impact domains (e.g., critical infrastructure, CBRNE), dual-use, and autonomous agents. \\
p3  & AI Safety Evaluation and Red-Teaming & Conduct safety evaluations and red-teaming to identify, assess, and mitigate risks of AI systems. \\
p4  & Responsible Information Sharing & Engage in responsible information sharing and incident reporting across organizations to reduce vulnerabilities and misuse of advanced AI systems. \\
p5  & Enabling and Fostering AI Safety Science & Promote R\&D on safety evaluation techniques to support institution design grounded in scientific knowledge. \\
p6  & Ensuring Security Throughout the AI Lifecycle & Invest in and implement robust security management across the AI lifecycle, including physical, cyber, and insider-threat controls. \\
p7  & Advocating for Policy and Governance Frameworks & Contribute to institutional design and policy/governance frameworks (e.g., certification) that improve AI safety while enabling innovation. \\
p8  & Ensuring Data Quality & Manage data quality to suppress harmful outputs and improve reliability. \\
p9  & Protecting Personal Data and Intellectual Property & Protect citizens' rights including personal data and intellectual property. \\
p10 & Ensuring Inclusive Access & Deliver AI benefits to all toward an inclusive society. \\
p11 & Ensuring Transparency & Ensure appropriate disclosure and transparency about AI systems to increase public trust. \\
p12 & Human Capital Investment and Education & Improve digital literacy and education based on human-centric values. \\
p13 & International Coordination and Cooperation & Aim for global safety through international coordination, including interoperability and joint testing. \\
p14 & Realizing Opportunities and Transformations & Realize business and social transformation via public/industry/government use and support for SMEs/startups. \\
p15 & Establishing Effective Governance & Establish, implement, and disclose AI governance and risk management policies based on a risk-based approach. \\
\hline
\end{tabular}
\caption{AMAIS activity categories (p1, \dots, p15) used as the shared crosswalk coordinate system.}
\label{tab:amais-p1-p15}
\end{table*}

\section{Experiments}
This section describes the setup and the results of our two experiments.
The goal of our experiments is to evaluate (i) stability, defined as how consistently different models assign similar crosswalk similarity scores for the same document pair and activity item, and (ii) validity, defined as how closely LLM-assigned similarity scores align with human similarity judgments on selected document pairs.

We summarize the results of the stability evaluation using three heatmaps: the mean similarity heatmap (Figure~\ref{fig:crosswalk-heatmap}), the standard deviation heatmap (Figure~\ref{fig:crosswalk-heatmap-std}), and the model-pair MAD heatmap (Figure~\ref{fig:meta-evaluation-mad}) described in Section~\ref{sec:crosswalk_results}.
Section~\ref{sec:eval_human} presents the results of the validity evaluation referencing human annotations.
\subsection{LLM-based Crosswalk}
\subsubsection{Setup}
\label{sec:experimental-setup}

We use ten publicly available policy documents on AI safety (denoted $A,B,\dots,J$) listed in Table~\ref{tab:documents}.
Among the steps described in Section~\ref{sec:method}, we treat the activity item extraction and mapping step for each document as precomputed by ChatGPT-5.2 and fixed in the evaluation, and focus on evaluating the subsequent steps that compare the two documents with the extracted activity items and mappings held fixed. Throughout our experiments, both the prompts, the taxonomies, and the resulting crosswalk outputs are written in Japanese.
In our experiments, we use nine document pairs $j \in \{(A,B),\dots,(A,J)\}$ by fixing Document~$A$ (UK AISI, ``Our Research Agenda'') and varying the second document across $B,\dots,J$.
For each $(j,p)$, the model generates a summary of Document~$A$ for $p$, a summary of its counterpart document for $p$, a brief comparison, and a similarity score on a 0 to 5 scale.
To examine the sensitivity of our method to the choice of LLMs, we used five LLMs in our experiments as listed in Table~\ref{tab:models}. 
All other settings (e.g., prompts and temperature) are fixed and identical across experiments.
\begin{table*}[t]
\centering
\small
\begin{tabular}{p{0.25\linewidth} p{0.55\linewidth} p{0.15\linewidth}}
\hline
\textbf{Extracted activity item (English)} & \textbf{Supporting excerpt} & \textbf{AMAIS aspect} \\
\hline
Hiring technical experts and partnering with researchers across government, academia, and industry (p.~4). &
``we have hired technical experts from top industry and academic labs, ... We have built partnerships with leading AI labs, research organisations, academia, and segments of the UK government'' &
$p12$ (Human Capital Investment and Education) \\
\hline
Distilling research findings into best practices, standards, and protocols for AI safety and security (p.~4). &
``International protocols: Working with key partners across government, we distil key research findings into best practices, standards, and protocols for AI safety and security and cohere model developers, deployers, and international actors around them.'' &
$p7$ (Advocating for Policy and Governance Frameworks) \\
\hline
\end{tabular}
\caption{Two example activity items extracted from the UK-AISI document ``Our Research Agenda'' \citep{metadata_uk_saisi_research_agenda}.}
\label{tab:uk_saisi_activity_examples}
\end{table*}

\begin{table*}[t]
\centering
\small
\begin{tabular}{rp{90truemm}p{45truemm}}
\hline
\textbf{Label} & \textbf{Title} & \textbf{Entity} \\
\hline
$A$ & Our Research Agenda & \citealp{metadata_uk_saisi_research_agenda} \\
$B$ & ASEAN Guide on AI Governance and Ethics & \citealp{metadata_asean_guide_on_ai_gov_and_ethics} \\
$C$ & CAISI Research Program at CIFAR 2025 Year in Review: Building Safe AI for Canadians & \citealp{metadata_ca_research_year_in_review} \\
$D$ & Safe, Secure, and Trustworthy Development and Use of Artificial Intelligence & \citealp{metadata_eo14110} \\
$E$ & America's Action Plan & \citealp{metadata_trump_ai_action_plan} \\
$F$ & The Singapore Consensus on Global AI Safety Research Priorities & \citealp{metadata_sg_consensus_on_ai_safety} \\
$G$ & Governing with Artificial Intelligence & \citealp{metadata_oecd_governing_with_ai} \\
$H$ & Framework Act on the Development of Artificial Intelligence and Establishment of Trust (English translated draft) & \citealp{metadata_kr_ai_basic_act} \\
$I$ & National status report of AI Safety in Japan 2024 & \citealp{metadata_j_aisi} \\
$J$ & National Strategic Review 2025 & \citealp{metadata_fr_national_strategy} \\
\hline
\end{tabular}
\caption{Document set used in the case study and the label assignment adopted in this paper.}
\label{tab:documents}
\end{table*}

\begin{table*}[t]
\centering
\small
\begin{tabular}{rll}
\hline
\textbf{Method} & \textbf{Model ID} & \textbf{Model name} \\
\hline
a & mistral.mistral-large-3-675b-instruct & Mistral Large 3 \\
b & qwen.qwen3-next-80b-a3b & Qwen3 Next 80B A3B \\
c & google.gemma-3-27b-it & Gemma 3 27B \\
d & openai.gpt-oss-120b-1\_0 & gpt-oss-120b \\
e & nvidia.nemotron-nano-3-30b & Nemotron Nano 3 30B \\
\hline
\end{tabular}
\caption{List of LLMs used in the experiments. All other settings (e.g., prompts and temperature) are fixed and identical across experiments.}
\label{tab:models}
\end{table*}

\paragraph{Mean similarity heatmap.}
For each $(j,p)$, we compute the average similarity score across models
\begin{equation}
\overline{s}_{j,p} = \frac{1}{|M|}\sum_{m\in M} s_{m,j,p}
\end{equation}
where $M$ denotes the set of LLMs $\{m_1, \dots, m_5\}.$

\paragraph{Standard deviation heatmap.}
For each $(j,p)$, we compute the standard deviation of $s_{m,j,p}$ across $m\in M$ to visualize how sensitive the results are to the choice of model:
\begin{equation}
\sigma_{j,p} = \sqrt{\frac{1}{|M|-1}\sum_{m\in M}\left(s_{m,j,p} - \overline{s}_{j,p}\right)^2}.
\end{equation}

\paragraph{Model pair mean absolute difference (MAD) heatmap.}
For each pair of ordered models $(m_1,m_2)$, we compute the mean absolute difference of scores, averaged over all pairs of documents and all aspects:
\begin{equation}
\mathrm{MAD}_{m_1,m_2} = \frac{1}{|J||P|}\sum_{j\in J}\sum_{p\in P} \left| s_{m_1,j,p} - s_{m_2,j,p} \right|.
\end{equation}
This aggregation summarizes how closely different models agree in their similarity scoring behavior.

\begin{figure*}[t]
  \centering
  \includegraphics[width=\linewidth]{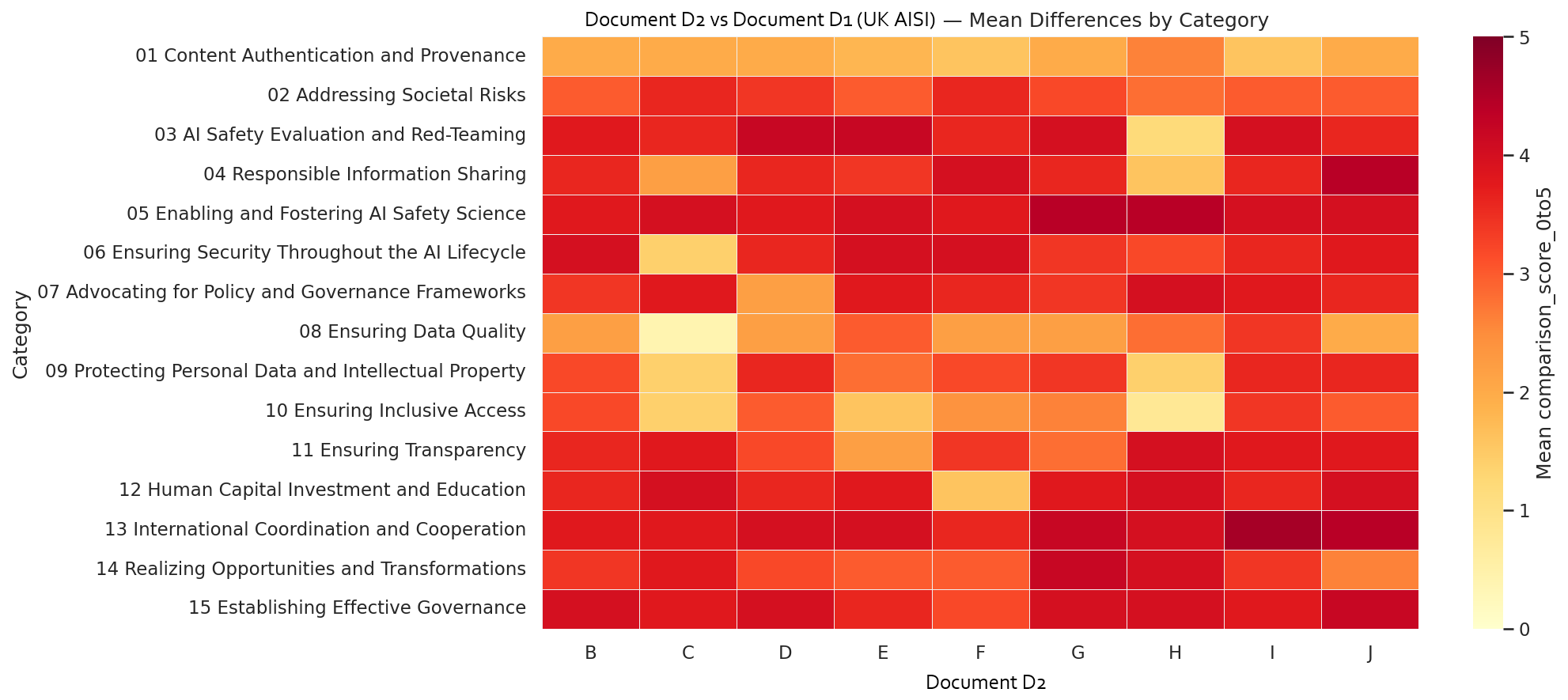}
  \caption{Heatmap of similarity scores where rows correspond to AI-safety activity items and columns correspond to Document~$D_2$ (Document~$D_1$ is fixed to UK-AISI). Each value is averaged over results from five models.}
  \label{fig:crosswalk-heatmap}
\end{figure*}

\begin{figure*}[t]
  \centering
  \includegraphics[width=\linewidth]{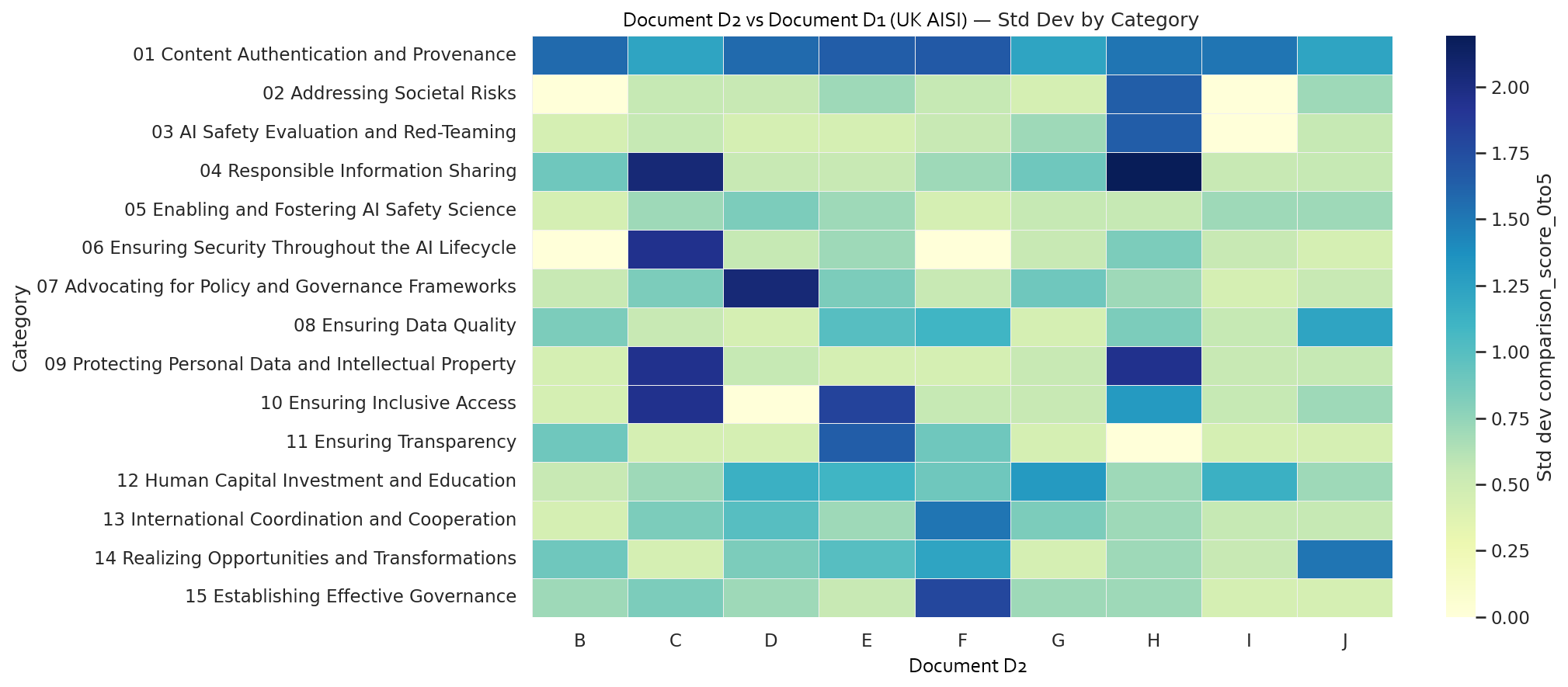}
  \caption{Heatmap of the standard deviation of similarity scores where rows correspond to AI-safety activity items and columns correspond to Document~$D_2$ (Document~$D_1$ is fixed to UK-AISI). Lower values indicate closer agreement among the results produced by five AI models.}
  \label{fig:crosswalk-heatmap-std}
\end{figure*}

\begin{figure}[t]
  \centering
  \includegraphics[width=\linewidth]{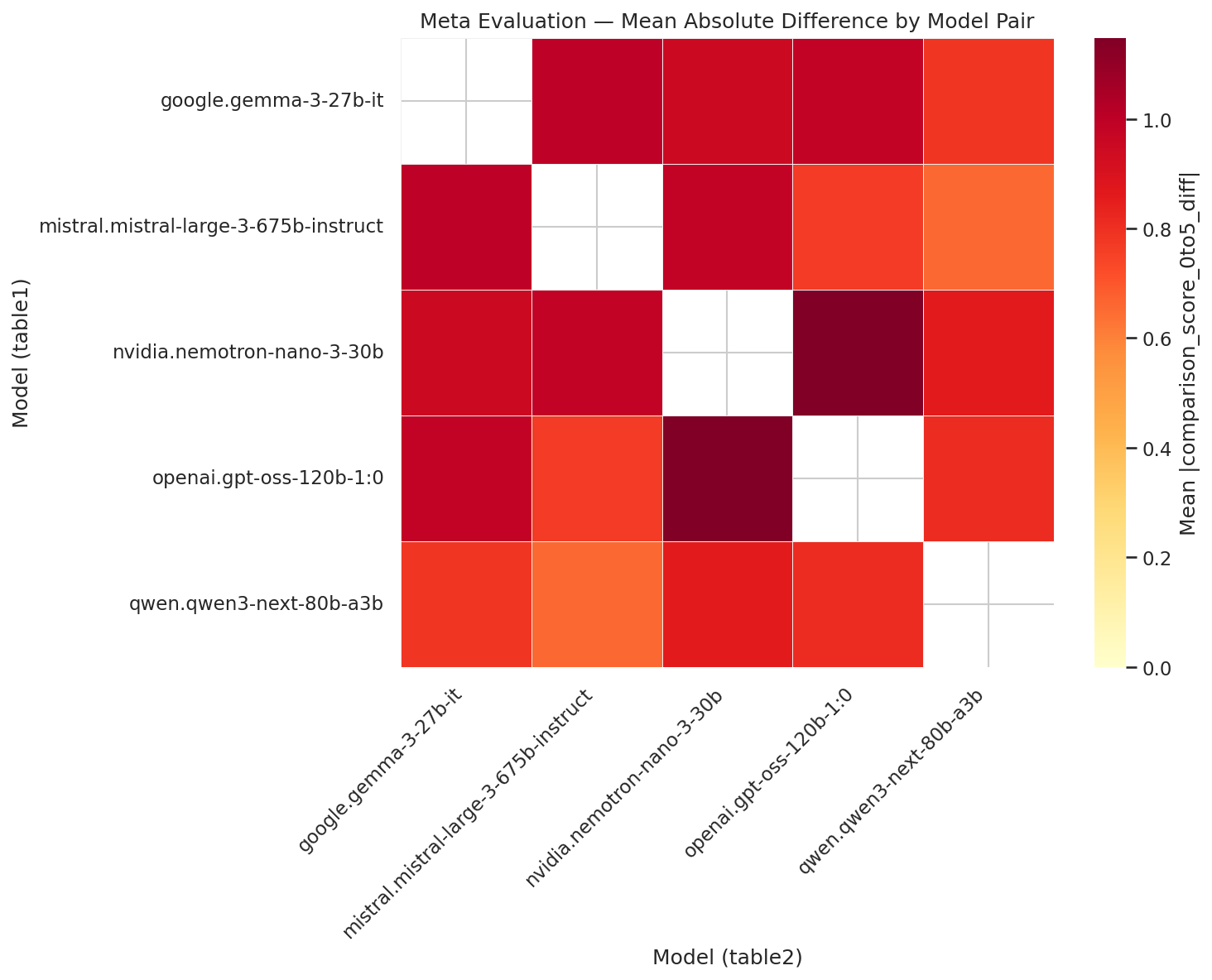}
  \caption{Heatmap comparing crosswalk results produced by pairs of AI models (out of five). Rows correspond to one model and columns to another. Each cell reports the mean absolute difference between the two models' crosswalk results, averaged over 9 document pairs (Document 1 fixed; Document 2 varying across nine documents) and 15 activity items.}
  \label{fig:meta-evaluation-mad}
\end{figure}
\subsubsection{Crosswalk Results}
\label{sec:crosswalk_results}
\paragraph{Average similarity across document pairs and activity items}
The mean similarity heatmap (Figure~\ref{fig:crosswalk-heatmap}) reports the average similarity score across the five models for each combination of a document-pair and an aspect. Higher values indicate closer alignment between Document~$D_1$ and the corresponding Document~$D_2$ for that activity item. We observe consistently low scores for $p1$ (Content Authentication and Provenance) across document pairs, whereas $p5$ (Enabling and Fostering AI Safety Science) and $p13$ (International Coordination and Cooperation) show comparatively high scores.

\paragraph{Across-model variability}
The standard deviation heatmap (Figure~\ref{fig:crosswalk-heatmap-std}) visualizes the variability of similarity scores across the five models for each combination of a document-pair and an aspect. We find large variability for $p1$, indicating that model judgments vary for this aspect. For the Document~$C$, several activity items (e.g., $p4$, $p6$, $p9$, $p10$) also show comparatively large variability, suggesting the stronger sensitivity to model choice for these settings.

\paragraph{Model-pair Disagreement Evaluation}
Figure~\ref{fig:meta-evaluation-mad} aggregates absolute differences in similarity scores between models. This indicates that Qwen3 and Gemma3 exhibit comparatively large disagreements, and that Nemotron tends to be relatively less aligned with other models.

\subsubsection{Implication}
The experimental results demonstrate that model selection is a critical hyperparameter in automated policy analysis, significantly influencing the output even when prompts and taxonomy are fixed. The high variability observed in specific categories, such as \textit{Content Authentication and Provenance} ($p1$), suggests that certain policy concepts may be more ambiguous or technically contested within the models' training data, leading to divergent interpretations. Furthermore, the distinct disagreement patterns between model families (e.g., the high disagreement between Qwen3/Gemma3 and the outlier behavior of Nemotron)  indicate that a single-model approach risks introducing model-specific biases into the crosswalk. Consequently, robust automated crosswalks should ideally employ model ensembling or majority-voting mechanisms rather than relying on a single LLM. Since inconsistent interpretations across models undermine the reliability of a shared coordinate system, ensemble methods are crucial to ensure the stability and neutrality required for policy interoperability tasks.

\subsection{Validation with Human Annotation}
\label{sec:eval_human}
\subsubsection{Setup}
We conducted a human evaluation to assess the validity of similarity scores assigned by LLMs by comparing them with human similarity judgments.

A crosswalk compares two documents by reorganizing their content under a shared set of aspects.
In this study, we use the 15 AMAIS activity categories, i.e., aspects, as the shared crosswalk taxonomy.
We evaluated two document pairs for human evaluation: $(A, D)$ and $(A, E)$ (See Table~\ref{tab:documents} for the document labels).

Three annotators (AI safety researchers) manually performed the crosswalk evaluation.
Annotators were provided with the extracted activity items grouped into 15 aspects and with the crosswalk task outputs (the Document~$A$ summary, the Document~$D/E$ summary, and the comparison result).
For each aspect, they assigned a similarity score using the following 6-level rubric (5: Almost identical, 4: Broadly consistent, 3: Partially consistent, 2: Limited consistency, 1: Slight consistency, 0: Almost different content).

Tables~\ref{tab:human-eval-eo14110} and~\ref{tab:human-eval-action-plan} report the similarity scores among the annotators for each aspect, together with the standard deviation and median among the three annotator scores.
Across the 15 aspects, we observe no large differences between annotators, and the standard deviations are generally within 1.
\begin{table*}[t]
\centering
\small
\setlength{\tabcolsep}{4pt}
\renewcommand{\arraystretch}{1.05}
\begin{tabularx}{\textwidth}{r X r r r r r}
\toprule
\textbf{ID} & \textbf{Activity item} & \textbf{Score 1} & \textbf{Score 2} & \textbf{Score 3} & \textbf{Std.\ dev.} & \textbf{Median} \\
\midrule
1 & Content Authentication and Provenance & 0 & 1 & 2 & 1.000 & 1 \\
2 & Addressing Societal Risks & 3 & 2 & 2 & 0.577 & 2 \\
3 & AI Safety Evaluation and Red-Teaming & 4 & 3 & 3 & 0.577 & 3 \\
4 & Responsible Information Sharing & 4 & 4 & 4 & 0.000 & 4 \\
5 & Enabling and Fostering AI Safety Science & 2 & 3 & 2 & 0.577 & 2 \\
6 & Ensuring Security Throughout the AI Lifecycle & 3 & 3 & 3 & 0.000 & 3 \\
7 & Advocating for Policy and Governance Frameworks & 0 & 1 & 1 & 0.577 & 1 \\
8 & Ensuring Data Quality & 0 & 2 & 2 & 1.155 & 2 \\
9 & Protecting Personal Data and Intellectual Property & 3 & 1 & 1 & 1.155 & 1 \\
10 & Ensuring Inclusive Access & 0 & 0 & 0 & 0.000 & 0 \\
11 & Ensuring Transparency & 0 & 0 & 0 & 0.000 & 0 \\
12 & Human Capital Investment and Education & 3 & 3 & 3 & 0.000 & 3 \\
13 & International Coordination and Cooperation & 1 & 2 & 1 & 0.577 & 1 \\
14 & Realizing Opportunities and Transformations & 3 & 2 & 2 & 0.577 & 2 \\
15 & Establishing Effective Governance & 1 & 2 & 1 & 0.577 & 1 \\
\bottomrule
\end{tabularx}
\caption{Human similarity scores for the crosswalk between Document~$A$ and Document~$D$. Std.\ dev.\ is computed across the three annotator scores; values are rounded to three decimals.}
\label{tab:human-eval-eo14110}
\end{table*}

\begin{table*}[t]
\centering
\small
\setlength{\tabcolsep}{4pt}
\renewcommand{\arraystretch}{1.05}
\begin{tabularx}{\textwidth}{r X r r r r r}
\toprule
\textbf{ID} & \textbf{Activity item} & \textbf{Score 1} & \textbf{Score 2} & \textbf{Score 3} & \textbf{Std.\ dev.} & \textbf{Median} \\
\midrule
1 & Content Authentication and Provenance & 0 & 2 & 2 & 1.155 & 2 \\
2 & Addressing Societal Risks & 3 & 3 & 3 & 0.000 & 3 \\
3 & AI Safety Evaluation and Red-Teaming & 5 & 4 & 3 & 1.000 & 4 \\
4 & Responsible Information Sharing & 4 & 3 & 3 & 0.577 & 3 \\
5 & Enabling and Fostering AI Safety Science & 2 & 3 & 2 & 0.577 & 2 \\
6 & Ensuring Security Throughout the AI Lifecycle & 0 & 2 & 3 & 1.528 & 2 \\
7 & Advocating for Policy and Governance Frameworks & 0 & 0 & 0 & 0.000 & 0 \\
8 & Ensuring Data Quality & 1 & 1 & 3 & 1.155 & 1 \\
9 & Protecting Personal Data and Intellectual Property & 3 & 3 & 2 & 0.577 & 3 \\
10 & Ensuring Inclusive Access & 3 & 4 & 4 & 0.577 & 4 \\
11 & Ensuring Transparency & 0 & 2 & 2 & 1.155 & 2 \\
12 & Human Capital Investment and Education & 3 & 5 & 3 & 1.155 & 3 \\
13 & International Coordination and Cooperation & 4 & 5 & 2 & 1.528 & 4 \\
14 & Realizing Opportunities and Transformations & 3 & 4 & 2 & 1.000 & 3 \\
15 & Establishing Effective Governance & 3 & 3 & 3 & 0.000 & 3 \\
\bottomrule
\end{tabularx}
\caption{Human similarity scores for the crosswalk between Document~$A$ and Document~$E$. Std.\ dev.\ is computed across the three annotator scores; values are rounded to three decimals.}
\label{tab:human-eval-action-plan}
\end{table*}

\subsubsection{Human Annotations vs.\ LLM Crosswalk Scores}
\label{sec:human-vs-llm}

We compare crosswalk similarity scores assigned by three human annotators with similarity scores produced by LLM-based crosswalk generation.
For each pair, the comparison is conducted for each of the 15 AMAIS aspects.

\begin{figure}[t]
\centering
\includegraphics[width=\linewidth]{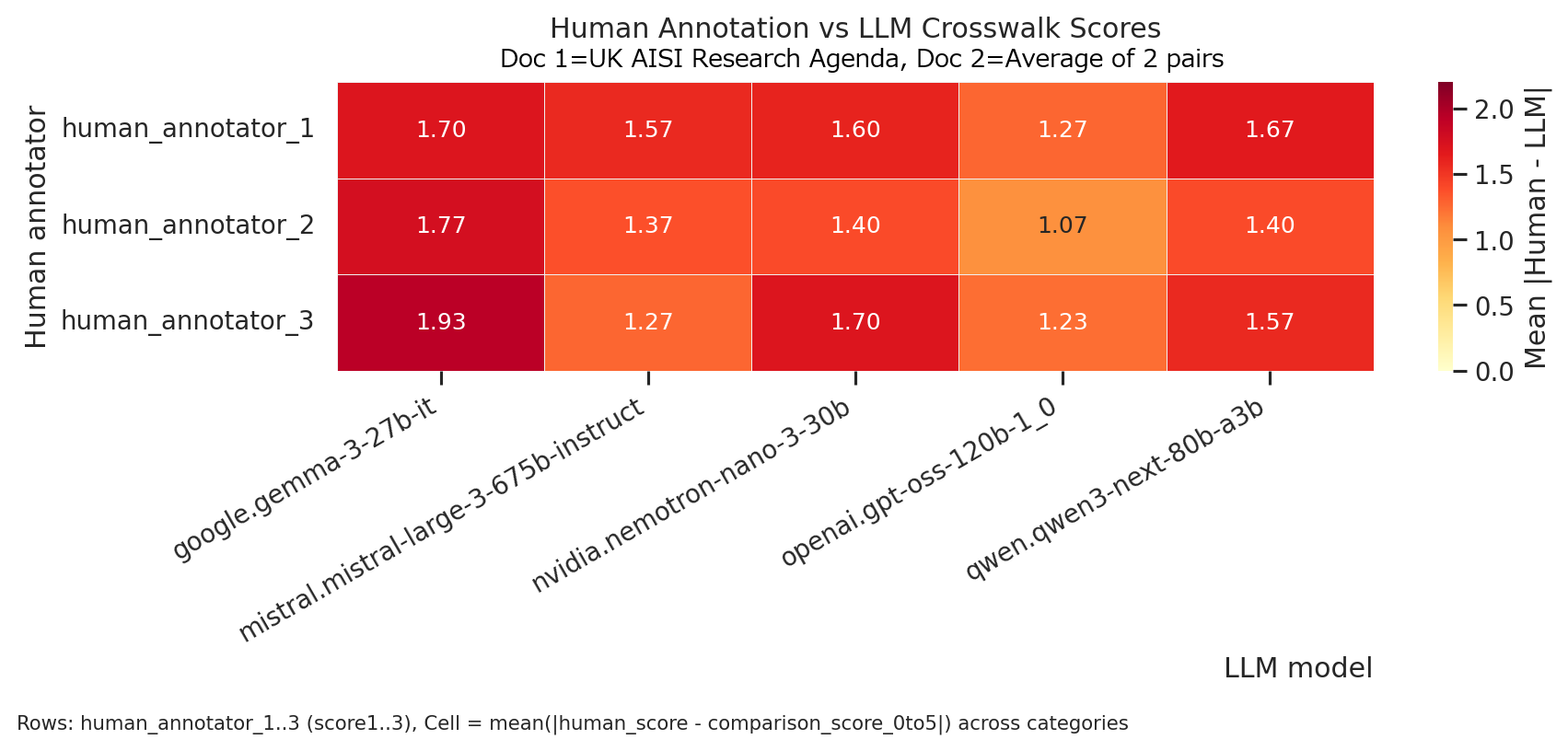}
\caption{Mean absolute difference between human annotator scores and LLM-produced crosswalk similarity scores, averaged over the 15 AMAIS activity items and the two document pairs of $(A, D)$ and $(A, E)$.}
\label{fig:human-vs-llm-mad-avg2pairs}
\end{figure}

\begin{figure}[t]
\centering
\includegraphics[width=\linewidth]{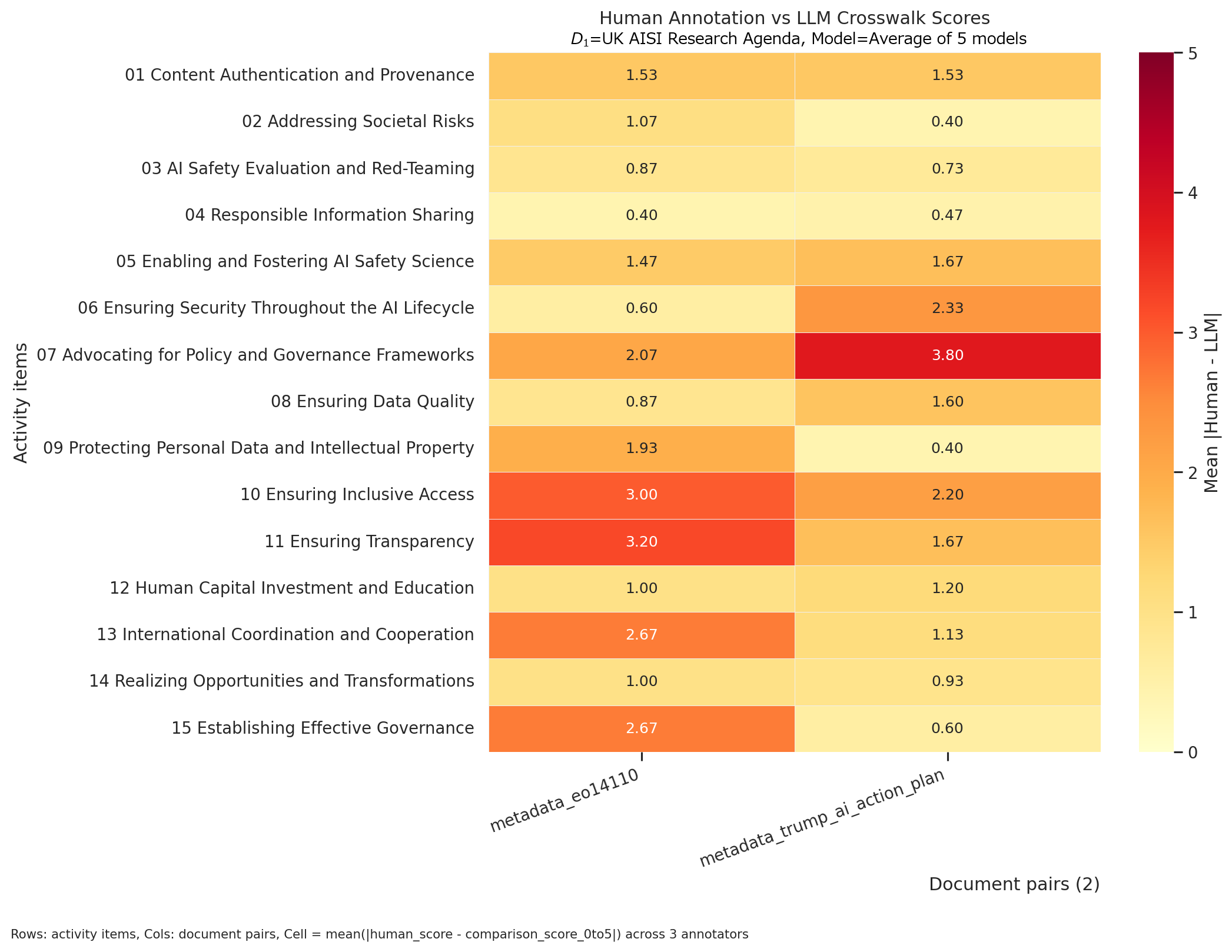}
\caption{Mean absolute difference between human annotator scores and LLM-produced crosswalk similarity scores, averaged over annotators for the two document pairs of $(A, D)$ and $(A, E)$.}
\label{fig:human-vs-llm-mad-avg-annotators}
\end{figure}

For each aspect, we compute the score difference between a human annotator and an LLM result, take the absolute value, and then average over the 15 aspects.
Figure~\ref{fig:human-vs-llm-mad-avg2pairs} aggregates the resulting mean absolute differences over the two document pairs. Figure~\ref{fig:human-vs-llm-mad-avg-annotators} aggregates the resulting mean absolute differences over annotators, highlighting the results between the different activity items and document pairs.

The mean absolute difference is between 1 and 2.
Across LLMs, the spread of mean absolute differences within each annotator is 0.43 (Annotator~1), 0.70 (Annotator~2), and 0.70 (Annotator~3).
Within a fixed model, the variation attributable to annotator differences is approximately within 0.3, indicating that annotator-to-annotator variation is smaller than model-to-model variation under this metric.

\subsubsection{Implication}
The comparison with human judgments reveals a calibration gap between LLM-generated scores and expert consensus. While human annotators exhibited high agreement (low standard deviation), the LLM scores deviated from human scores with a Mean Absolute Difference (MAD) between 1.0 and 2.0 on a 5-point scale. This suggests that while the task itself is well-defined for experts, LLMs currently struggle to align their scoring magnitude with human intuition, likely due to the lack of few-shot examples or detailed scoring rubrics in the prompt. Therefore, in its current state, the framework is best utilized as a human-in-the-loop assistive tool—generating draft summaries and identifying potential diffs for human review—rather than a fully autonomous evaluation system. Future work must focus on value alignment and prompt calibration to bridge the quantitative gap between algorithmic and human policy assessment. Establishing such alignment is a prerequisite for the proposed framework to serve as a trusted, interoperable standard across diverse stakeholders.

\section{Conclusion}
\label{sec:conclusion}
We presented a framework for automated crosswalk analysis of AI safety policy document pairs. By grounding crosswalks in AMAIS activity categories (p1, \dots, p15) and defining a structured LLM task that produces per-aspect summaries, diffs, and a similarity score, the framework provides a reusable coordinate system for interoperability-oriented comparisons. A case study with ten documents and five AI models, summarized via aggregate heatmaps, qualitatively indicates that some document pairs and activity items yield lower similarity and higher across-model variability, and that model choice affects crosswalk results. 

\section{Limitations}
\label{sec:limitations}
\begin{itemize}
  \item \textbf{Precomputed extraction and mapping:} As described in Section~\ref{sec:experimental-setup}, we treat activity item extraction and the mapping to aspects as precomputed and fixed. As a result, this study does not evaluate errors that may arise in these steps, such as hallucinated activity items or ambiguity in aspect assignment. Future work should develop and evaluate a full pipeline that includes extraction and mapping, and should assess how these steps affect the resulting crosswalk outputs.
  \item \textbf{Language setting:} As described in Section~\ref{sec:experimental-setup}, we used Japanese for the prompts, the taxonomy descriptions, and the generated crosswalk outputs. We therefore do not analyze differences that may arise when the same procedure is conducted in other languages. Future work should evaluate the framework in multiple languages and examine how language choice affects the crosswalk results.
  \item \textbf{Aspect definition ambiguity/overlap:} Some activity categories may be interpreted differently across models, and categories may overlap in practice, leading to the observed variability.
  \item \textbf{Score definition and prompt calibration:} While we specified a 0-5 similarity scale, detailed scoring rubrics and few-shot examples were not provided in the LLM prompts. This likely contributed to the calibration gap between LLM and human scores.
  \item \textbf{Human evaluation scale:} We conducted a validity evaluation with human experts; however, the scale was limited to three annotators and two document pairs. Future work requires larger-scale human evaluation with rigorous inter-annotator agreement analysis. 
  \item \textbf{Scope of documents and pairing strategy:} The dataset contains ten documents and fixes Document~$D_1$ to UK-AISI, yielding nine pairs. This design may not generalize to other anchor documents or fully cross-paired analyses. 
\end{itemize}

\section{Bibliographical References}
\bibliographystyle{lrec2026-natbib}
\bibliography{policy_docs,related_work}

\onecolumn
\begin{paracol}{2}
\section{Appendices}
\subsection{AMAIS Activity - Activity Extraction Prompt}

In this section we show an English translation of the prompt used for AMAIS activity extraction. The XML structure and schema field names are preserved from the original operational prompt, while the Japanese prose is translated into English for readability in the manuscript.

\switchcolumn
\stepcounter{section}
\stepcounter{subsection}
\subsection{AMAIS Activity - Difference Analysis Prompt}
\label{sec:amais-prompt-template}
In this section we show an English translation of the prompt used for AMAIS activity-difference analysis. The XML structure and schema field names are preserved from the original operational prompt, while the Japanese prose is translated into English for readability in the manuscript.
\end{paracol}

\begin{lstlisting}[caption=AMAIS Activity-Activity Extraction Prompt,basicstyle=\ttfamily\small,breaklines=true,breakatwhitespace=false,columns=fullflexible]
<task>
  <role>
    You are a highly precise text analysis assistant.
    Analyze the provided document and extract information related to each of the 15 activities defined in the Activity Map on AI Safety (AMAIS). 
  </role>

  <documents>
    <!-- Insert the source text to be analyzed here, or provide separate file-loading instructions. -->
    <document id="doc-1">
      <![CDATA[
      (Paste the full text to analyze here)
      ]]>
    </document>
  </documents>

  <instructions>
    <step number="1">
      <objective>Extract activities described in the document. An activity is a document-backed, actionable unit of work that designates a specific actor to perform a clearly defined action on a specified object or topic.</objective>
      <requirements>
        <field>title</field>
        <field>description</field>
        <field>page_number</field>
        <field>excerpts</field>
      </requirements>
      <notes>
        <note>Use page_number as integer when available; if absent, infer from closest heading or section marker, else set "unknown".</note>
        <note>excerpts should be verbatim quotes (up to 2--3 sentences) supporting your extraction.</note>
      <note>Regarding activity extraction, you may extract more than the 15 activity items defined by AMAIS. Treat any items with different actors or actions, or with different deliverables, as separate activities.</note>
      </notes>
    </step>

    <step number="2">
      <objective>For each extracted activity, perform classification and evaluation.</objective>
      <substep number="2.1">
        <action>Map the activity to exactly one AMAIS category using the <AMAIS_categories> reference below.</action>
      </substep>
      <substep number="2.2">
        <action>Assign an extent score from 1 to 5.</action>
        <scale>
          <value number="1">Negative extent (e.g., stop, strong opposition)</value>
          <value number="2">Cautious/limiting stance</value>
          <value number="3">Neutral/ambivalent or exploratory</value>
          <value number="4">Supportive/enable with safeguards</value>
          <value number="5">Strongly positive (start, encourage, promote)</value>
        </scale>
      </substep>
	<substep number="2.3">
	        <action>Assign a confidence of your reasoning from 0.0 to 1.0</action>
	      </substep>
      <substep number="2.4">
        <action>Provide reasoning for both the category mapping (2.1), the extent score (2.2) and the confidence (2.3). Cite the excerpt(s) that justify your choices.</action>
      </substep>
    </step>

    <quality_checks>
      <check>Every activity must have a single mapped_category (no multi-labels).</check>
      <check>Reasoning must reference at least one excerpt.</check>
      <check>If confidence in mapping &lt; 0.6, add &lt;ambiguous true="yes"/&gt; and propose the next best category in &lt;alternative_category/&gt;.</check>
    </quality_checks>
  </instructions>

  <AMAIS_categories>
    <!-- Each category element includes the English name and ID attributes plus the three required child elements. -->
    <category id="1" name="Content Authentication and Provenance">
      <category_name_jp>Content Authentication and Provenance Mechanisms</category_name_jp>
      <description>Where technically feasible, develop and deploy reliable authentication and provenance mechanisms, such as watermarking and related techniques, so that users can identify AI-generated content.</description>
      <keywords>Originator Profile; Disinformation; Hallucination; Watermarking; Synthetic Contents; Provenance mechanisms; Disclaimer; AI Label</keywords>
    </category>

    <category id="2" name="Addressing Societal Risks">
      <category_name_jp>Addressing Societal Risks</category_name_jp>
      <description>Take appropriate measures to address risks in areas with major impacts on human life and society, including critical infrastructure, CBRNE, dual-use concerns, and autonomous agents.</description>
      <keywords>Dual-use; Foundation Model; AGI; AI-agent; GPAI; Risk management for CBRN; AI for Critical Infrastructure; IT/OT; Cognitive and Behavioral Manipulation; Profiling; Job Market in the age of AI</keywords>
    </category>

    <category id="3" name="AI Safety Evaluation and Red-Teaming">
      <category_name_jp>AI Safety Evaluation and Red-Teaming</category_name_jp>
      <description>Conduct safety evaluations and red-teaming to identify, assess, and mitigate risks in AI systems.</description>
      <keywords>Threat Actor Uplift Evaluation; External Testing; Automated Evaluation; Test-bed; Robustness; Alignment</keywords>
    </category>

    <category id="4" name="Responsible Information Sharing">
      <category_name_jp>Responsible Information Sharing</category_name_jp>
      <description>Promote responsible information sharing and incident reporting across organizations to reduce vulnerabilities and misuse of advanced AI systems.</description>
      <keywords>Bounty Program; Multi-Stakeholder; Incident Response and Sharing among Industry, Academia and Government; Early Warning Information Sharing; Incident Report</keywords>
    </category>

    <category id="5" name="Enabling and Fostering AI Safety Science">
      <category_name_jp>Enabling and Fostering AI Safety Science</category_name_jp>
      <description>Advance research and development of safety evaluation technologies to support institution design grounded in scientific knowledge.</description>
      <keywords>Academic Research; Grants and Startups by Government; Safety for Emerging Technology; Foundation Model</keywords>
    </category>

    <category id="6" name="Ensuring Security Throughout the AI Lifecycle">
      <category_name_jp>Ensuring Security Throughout the AI Lifecycle</category_name_jp>
      <description>Invest in and implement strong security management across the AI lifecycle, including physical, cyber, and insider-threat protections.</description>
      <keywords>Cyber; Physical Access Control; Information Security; Risk Mitigation; Internal Threat Detection Program; Security for AI; AI for Security</keywords>
    </category>

    <category id="7" name="Advocating for Policy and Governance Frameworks">
      <category_name_jp>Advocating for Policy and Governance Frameworks</category_name_jp>
      <description>Contribute to institution design, including certification systems and related measures, that supports the maintenance and improvement of AI safety while promoting innovation.</description>
      <keywords>Developing Guidelines; Identifying Value-chain; Addressing AI Safety Washing; Ensuring Fair Competition; Certification System; Taxonomy and Terminology</keywords>
    </category>

    <category id="8" name="Ensuring Data Quality">
      <category_name_jp>Ensuring Data Quality</category_name_jp>
      <description>Manage data quality to suppress harmful outputs and improve reliability.</description>
      <keywords>Traceability; Output Attribution; Enhancing Interpretability</keywords>
    </category>

    <category id="9" name="Protecting Personal Data and Intellectual Property">
      <category_name_jp>Protecting Personal Data and Intellectual Property</category_name_jp>
      <description>Protect citizens' rights, including personal data and intellectual property.</description>
      <keywords>Privacy; Copyright; Safeguard</keywords>
    </category>

    <category id="10" name="Ensuring Inclusive Access">
      <category_name_jp>Ensuring Inclusive Access</category_name_jp>
      <description>Deliver the benefits of AI to everyone in pursuit of a society where no one is left behind.</description>
      <keywords>Accessibility; Safety Net; Diversity; Outreach; Human Welfare; Protection from Disasters</keywords>
    </category>

    <category id="11" name="Ensuring Transparency">
      <category_name_jp>Ensuring Transparency</category_name_jp>
      <description>Ensure appropriate disclosure and transparency regarding AI systems to strengthen trust among citizens and society.</description>
      <keywords>Responsible AI Development; Ethics; Trustworthiness; Accountability; Fairness; Transparency Report; Model Card; System Card; Data Card; Human-Centric</keywords>
    </category>

    <category id="12" name="Human Capital Investment and Education">
      <category_name_jp>Human Capital Investment and Education</category_name_jp>
      <description>Provide education and improve digital literacy based on human-centered values.</description>
      <keywords>Outreach; Certification System; School Education</keywords>
    </category>

    <category id="13" name="International Coordination and Cooperation">
      <category_name_jp>International Coordination and Cooperation</category_name_jp>
      <description>Pursue global safety through international coordination, including interoperability and joint testing.</description>
      <keywords>Interoperability; Guardrail; Standards Development Organizations; Cross-border; Joint Testing; Cross-disciplinary; Scientific</keywords>
    </category>

    <category id="14" name="Realizing Opportunities and Transformations">
      <category_name_jp>Realizing Opportunities and Transformations</category_name_jp>
      <description>Realize business and societal transformation through public-sector, industrial, and governmental use, as well as support for SMEs and startups.</description>
      <keywords>Public Sector; Manufacturing; Robotics and Mobility Logistics and Healthcare; Government; SMEs and Startups</keywords>
    </category>

    <category id="15" name="Establishing Effective Governance">
      <category_name_jp>Establishing Effective Governance</category_name_jp>
      <description>Formulate, implement, and disclose AI governance and risk-management policies based on a risk-based approach.</description>
      <keywords>Risk Management; Management System; Risk Assessment; Accountability</keywords>
    </category>
  </AMAIS_categories>

  <output_format>
    <!-- The model should return an array in this format. -->
    <activities>
      <activity>
        <title></title>
        <description></description>
        <page_number></page_number>
        <excerpts></excerpts>
        <mapped_category id="1" name="Content Authentication and Provenance"/>
        <extent_score>1-5</extent_score>
        <confidence>0.0-1.0</confidence>
        <reasoning></reasoning>
        <!-- Only include the following when the case is ambiguous. -->
        <ambiguous true="no"/>
        <alternative_category id="" name=""/>
      </activity>
      <!-- repeat for each extracted activity -->
    </activities>
  </output_format>

  <disambiguation_rules>
    <rule>First, try keyword overlap with <keywords>. If multiple categories match, prefer the one whose description semantically aligns with the excerpt.</rule>
    <rule>If still tied, inspect surrounding sentences for policy/tech/security context to break ties (e.g., "incident report" \rightarrow category 4).</rule>
    <rule>When in doubt, set ambiguous="yes", propose alternative_category, and lower confidence.</rule>
  </disambiguation_rules>

  <final_checks>
    <check>Total activities extracted \geq number of clearly distinct initiatives mentioned in the document.</check>
    <check>No activity lacks excerpts or reasoning.</check>
   <check> confidence is between 0.0 and 1.0</check>
    <check>extent_score is integer 1--5 only.</check>
  </final_checks>
</task>
\end{lstlisting}

\begin{lstlisting}[caption=AMAIS Activity-Difference Analysis Prompt]
<POML version="1.0">
  <meta>
    <title>AMAIS Activity-Difference Analysis Prompt</title>
    <author>IPA/Analysis Support</author>
    <language>en-US</language>
    <style>declarative style</style>
    <purpose>Extract and compare differences in initiatives across the 15 AMAIS activity categories from the metadata (XML) of two documents and generate JSON.</purpose>
  </meta>

  <!-- Input: XML for each document conforming to input_format (activities array) -->
  <inputs>
    <input id="document_A_xml" format="xml">
      <title></title>
      <![CDATA[
{{DOCUMENT_A_XML}}
      ]]>
    </input>
    <input id="document_B_xml" format="xml">
      <title></title>
      <![CDATA[
{{DOCUMENT_B_XML}}
      ]]>
    </input>
    <!-- AMAIS category definitions (must not be changed): fix them inside the prompt and treat id and name as canonical -->
    <input id="AMAIS_categories" format="xml">

      <![CDATA[
<AMAIS_categories>
  <category id="1"  name="Content Authentication and Provenance">
    <category_name_jp>Content authentication and provenance mechanisms</category_name_jp>
    <description>When technically feasible, develop and deploy reliable authentication and provenance mechanisms, such as digital watermarking and related techniques, so that users can identify AI-generated content.</description>
    <keywords>Originator Profile; Disinformation; Hallucination; Watermarking; Synthetic Contents; Provenance mechanisms; Disclaimer; AI Label</keywords>
  </category>
  <category id="2"  name="Addressing Societal Risks">
    <category_name_jp>Measures for societal risks</category_name_jp>
    <description>Appropriately address risks related to dual-use, autonomous agents, critical infrastructure, CBRNE, and other domains with major impacts on human life and society.</description>
    <keywords>Dual-use; Foundation Model; AGI; AI-agent; GPAI; Risk management for CBRN; AI for Critical Infrastructure; IT/OT; Cognitive and Behavioral Manipulation; Profiling; Job Market in the age of AI</keywords>
  </category>
  <category id="3"  name="AI Safety Evaluation and Red-Teaming">
    <category_name_jp>AI safety evaluation and red-teaming</category_name_jp>
    <description>Conduct safety evaluations and red-teaming to identify, assess, and mitigate risks in AI systems.</description>
    <keywords>Threat Actor Uplift Evaluation; External Testing; Automated Evaluation; Test-bed; Robustness; Alignment</keywords>
  </category>
  <category id="4"  name="Responsible Information Sharing">
    <category_name_jp>Responsible information sharing</category_name_jp>
    <description>Engage in responsible information sharing and incident reporting across organizations to mitigate vulnerabilities and misuse cases in advanced AI systems.</description>
    <keywords>Bounty Program; Multi-Stakeholder; Incident Response and Sharing among Industry, Academia and Government; Early Warning Information Sharing; Incident Report</keywords>
  </category>
  <category id="5"  name="Enabling and Fostering AI Safety Science">
    <category_name_jp>Promotion of AI safety science</category_name_jp>
    <description>Promote research and development of safety-evaluation technologies to support institution design grounded in scientific knowledge.</description>
    <keywords>Academic Research; Grants and Startups by Government; Safety for Emerging Technology; Foundation Model</keywords>
  </category>
  <category id="6"  name="Ensuring Security Throughout the AI Lifecycle">
    <category_name_jp>Ensuring security across the AI lifecycle</category_name_jp>
    <description>Invest in and implement robust security management across the entire AI lifecycle, including physical, cyber, and insider-threat countermeasures.</description>
    <keywords>Cyber; Physical Access Control; Information Security; Risk Mitigation; Internal Threat Detection Program; Security for AI; AI for Security</keywords>
  </category>
  <category id="7"  name="Advocating for Policy and Governance Frameworks">
    <category_name_jp>Promotion of policy and governance frameworks</category_name_jp>
    <description>Contribute to institution design that promotes innovation, including certification schemes and related mechanisms that help maintain and improve AI safety.</description>
    <keywords>Developing Guidelines; Identifying Value-chain; Addressing AI Safety Washing; Ensuring Fair Competition; Certification System; Taxonomy and Terminology</keywords>
  </category>
  <category id="8"  name="Ensuring Data Quality">
    <category_name_jp>Ensuring data quality</category_name_jp>
    <description>Manage data quality to suppress harmful outputs and improve reliability.</description>
    <keywords>Traceability; Output Attribution; Enhancing Interpretability</keywords>
  </category>
  <category id="9"  name="Protecting Personal Data and Intellectual Property">
    <category_name_jp>Protection of personal data and intellectual property</category_name_jp>
    <description>Protect citizens' rights, including personal information and intellectual property.</description>
    <keywords>Privacy; Copyright; Safeguard</keywords>
  </category>
  <category id="10" name="Ensuring Inclusive Access">
    <category_name_jp>Ensuring inclusive access</category_name_jp>
    <description>Deliver the benefits of AI to everyone in pursuit of a society in which no one is left behind.</description>
    <keywords>Accessibility; Safety Net; Diversity; Outreach; Human Welfare; Protection from Disasters</keywords>
  </category>
  <category id="11" name="Ensuring Transparency">
    <category_name_jp>Ensuring transparency</category_name_jp>
    <description>Ensure appropriate disclosure and transparency about AI systems in order to strengthen trust among citizens and society.</description>
    <keywords>Responsible AI Development; Ethics; Trustworthiness; Accountability; Fairness; Transparency Report; Model Card; System Card; Data Card; Human-Centric</keywords>
  </category>
  <category id="12" name="Human Capital Investment and Education">
    <category_name_jp>Human capital development and education</category_name_jp>
    <description>Improve digital literacy and provide education based on human-centered values.</description>
    <keywords>Outreach; Certification System; School Education</keywords>
  </category>
  <category id="13" name="International Coordination and Cooperation">
    <category_name_jp>International coordination and cooperation</category_name_jp>
    <description>Aim to ensure global safety through international coordination, including interoperability and joint testing.</description>
    <keywords>Interoperability; Guardrail; Standards Development Organizations; Cross-border; Joint Testing; Cross-disciplinary; Scientific</keywords>
  </category>
  <category id="14" name="Realizing Opportunities and Transformations">
    <category_name_jp>Realizing opportunities and transformations</category_name_jp>
    <description>Realize business and societal transformation through public-sector, industrial, and government use, as well as support for SMEs and startups.</description>
    <keywords>Public Sector; Manufacturing; Robotics and Mobility Logistics and Healthcare; Government; SMEs and Startups</keywords>
  </category>
  <category id="15" name="Establishing Effective Governance">
    <category_name_jp>Establishing governance</category_name_jp>
    <description>Formulate, implement, and disclose AI governance and risk-management policies based on a risk-based approach.</description>
    <keywords>Risk Management; Management System; Risk Assessment; Accountability</keywords>
  </category>
</AMAIS_categories>
      ]]>
    </input>
  </inputs>

  <!-- Output specification -->
  <outputs>
    <output id="json_table" format="application/json">
      <description>Dictionary-type JSON using category IDs as keys. For each category, include summaries for documents A and B, comparative findings, scores, representative values by document, differences, and raw values.</description>
      <filename>amais_diff_table.json</filename>
    </output>
  </outputs>

  <!-- Analysis rules -->
  <instructions>
    <rule>Assume that document_A_xml and document_B_xml both follow the format &lt;activities&gt;&lt;activity&gt;...&lt;/activity&gt;&lt;/activities&gt;.</rule>
    <rule>Normalize categories using the id (1-15) in &lt;mapped_category&gt; as the primary key. Even if the name varies, prioritize the id.</rule>
    <rule>For each category, collect the activity groups for A and B respectively and summarize the content (title, description, page_number, excerpts, extent_score, confidence, reasoning, ambiguous, alternative_category) in 1-2 sentences (roughly 100 Japanese characters in the original prompt).</rule>
    <rule>If multiple activities exist in the same category, summarize and describe the key points for both A and B.</rule>
    <rule>For extent_score, compute a representative value for documents A and B respectively, using weighted average (with confidence as the weight) when possible, otherwise simple average, or the single value if only one exists. In the JSON, store them as extent_docA and extent_docB, and compute extent_delta=extent_docA-extent_docB (if either is null, extent_delta must also be null).</rule>
    <rule>For confidence, compute representative values (avg/max) for documents A and B respectively. In the JSON, store confidence_docA and confidence_docB, and store only the numeric difference in averages as confidence_delta (e.g., confidence_delta=confidence_docA.avg-confidence_docB.avg).</rule>
    <rule>If ambiguous="yes" is included, explicitly note the uncertainty; if alternative_category is suggested, mention it in a footnote-like manner.</rule>
    <rule>If a category lacks any activity, assign the unknown label and set extent_score to 0.</rule>
    <rule>Comparison perspectives: describe the presence or absence of initiatives, the level of specificity, coverage (comprehensiveness), maturity (based on extent_score), evidence strength (confidence, excerpts, and page_number), and differences in direction.</rule>
    <rule>comparison_results must include a similarity score (integer 0-5) and a short explanation. Score definitions: 5=almost identical, 4=largely aligned, 3=partially aligned, 2=limited alignment, 1=slight alignment, 0=almost entirely different.</rule>
    <rule>If both sides have no activity in the category, explicitly state "Not applicable."</rule>
    <rule>Use Japanese terminology and unify the tone in the original prompt to the declarative style.</rule>
  </instructions>

  <!-- Procedure (implementation algorithm instructions) -->
  <procedure>
    <step>Parse document_A_xml and document_B_xml as XML and extract the activities arrays.</step>
    <step>Traverse category IDs 1 to 15 in AMAIS_categories in ascending order.</step>
    <step>For each category ID:
      <substep>A side: collect all activities whose mapped_category/@id matches the category ID and create a summary (up to 200 Japanese characters in the original prompt). Add representative page_number values and short excerpts (up to 1-2 items, each within 60 Japanese characters), and also compute representative values for extent_score and confidence for later A/B difference calculation.</substep>
      <substep>B side: create the summary in the same way.</substep>
      <substep>Write comparative findings (up to 200 Japanese characters in the original prompt). The perspectives are presence/absence, specificity, maturity, evidence strength, and direction.</substep>
    </step>
    <step>JSON output: a dictionary using category_id as the key. Include the following in each category. (The final output must be JSON only. Do not add headers, footers, or any extra explanation.)
      <substep>category_name_en, category_name_jp</substep>
      <substep>docA_summary, docB_summary, comparison_results (about 100 Japanese characters each in the original prompt)</substep>
      <substep>comparison_score_0to5</substep>
      <substep>unknown (true/false)</substep>
      <substep>extent_docA, extent_docB, extent_delta (extent_docA - extent_docB)</substep>
      <substep>confidence_docA (avg/max), confidence_docB (avg/max), confidence_delta (confidence_docA_avg - confidence_docB_avg)</substep>
      <substep>extent_raw_docA/extent_raw_docB (arrays), confidence_raw_docA/confidence_raw_docB (arrays)</substep>
      <substep>evidence_docA, evidence_docB (page_number, excerpts)</substep>
      <substep>notes (ambiguous, alternative_category). Always include both keys; if not applicable, use the empty string.</substep>
    </step>
  </procedure>

  <!-- Strict output-format specification -->
  <validation>
    <json id="json_table">
      <rules>
        <rule>The top level must be a dictionary keyed by category_id ("1" to "15").</rule>
        <rule>The output must be JSON only, with no explanatory text, headings, code fences, footers, or any additional prose before or after it.</rule>
        <rule>If unknown=true, comparison_score_0to5 must be 0.</rule>
        <rule>extent_raw_docA/B and confidence_raw_docA/B must store per-activity raw values separately for each document.</rule>
        <rule>notes.ambiguous and notes.alternative_category are required. If not applicable, use the empty string.</rule>
        <rule>confidence_delta must be either a numeric value (difference in averages) or null.</rule>
      </rules>
      <example>
        <![CDATA[
{
  "1": {
    "category_name_en": "Content Authentication and Provenance",
    "category_name_jp": "Content authentication and provenance mechanisms",
    "docA_summary": "Describes a policy of establishing authentication and provenance mechanisms, such as watermarking and digital signatures, to support identification of AI-generated content.",
    "docB_summary": "No corresponding activity can be found.",
    "comparison_results": "Comparison is not possible because no activity in the same category exists on the B side.",
    "comparison_score_0to5": 0,
    "unknown": true,
    "extent_docA": 4.0,
    "extent_docB": 0,
    "extent_delta": 4,
    "confidence_docA": {
      "avg": 0.72,
      "max": 0.9
    },
    "confidence_docB": null,
    "confidence_delta": null,
    "extent_raw_docA": [
      4.0
    ],
    "extent_raw_docB": [],
    "confidence_raw_docA": [
      0.9
    ],
    "confidence_raw_docB": [],
    "evidence_docA": {
      "page_number": [
        "12"
      ],
      "excerpts": [
        "It adopted a mission statement together with 'Track 1: Mitigating the Risks of Synthetic Content.'"
      ]
    },
    "evidence_docB": null,
    "notes": {
      "ambiguous": "yes",
      "alternative_category": "11 Ensuring Transparency"
    }
  },
  "2": {
    "category_name_en": "Addressing Societal Risks",
    "category_name_jp": "Measures for societal risks",
    "docA_summary": "Promotes efforts to assess the risk that AI may be used for cyberattacks and to build capability definitions and a risk-assessment framework.",
    "docB_summary": "Presents a policy for designing and implementing a risk-based governance structure, including escalation paths for high-risk AI and the establishment of ethics committees.",
    "comparison_results": "A focuses on cyber-capability assessment, whereas B emphasizes risk-based governance design. Their directions differ, but both aim to reduce societal risks.",
    "comparison_score_0to5": 3,
    "unknown": false,
    "extent_docA": 4.0,
    "extent_docB": 4.0,
    "extent_delta": 0.0,
    "confidence_docA": {
      "avg": 0.73,
      "max": 0.86
    },
    "confidence_docB": {
      "avg": 0.9,
      "max": 0.9
    },
    "confidence_delta": -0.17,
    "extent_raw_docA": [
      4.0,
      5.0
    ],
    "extent_raw_docB": [
      4.0,
      4.0,
      5.0
    ],
    "confidence_raw_docA": [
      0.73,
      0.86
    ],
    "confidence_raw_docB": [
      0.9,
      0.88,
      0.9
    ],
    "evidence_docA": {
      "page_number": [
        "8"
      ],
      "excerpts": [
        "We define the cyber capabilities AI models would need to have to facilitate these risk scenarios, across a range of cyber domains."
      ]
    },
    "evidence_docB": {
      "page_number": [
        "18",
        "19",
        "63"
      ],
      "excerpts": [
        "Deployers can also consider setting up a multi-disciplinary, central governing body, such as an AI Ethics Advisory Board or Ethics Committee, to oversee AI governance efforts, provide independent advice, and develop standards, guidelines, tools, and templates to help other teams design, develop, and deploy AI responsibly.",
        "Internal governance structures can also be designed for escalation of ethical issues, where AI systems and use cases that are of higher risk are escalated to a governing body with higher authority for review and decision-making.",
        "Consider defining separate roles and responsibilities for business and technical staff... Technical staff responsible for data practices, security, stability, error handling"
      ]
    },
    "notes": {
      "ambiguous": "no",
      "alternative_category": ""
    }
  },
        ]]>
      </example>
    </json>
  </validation>

  <!-- Generation command: always return the JSON artifact -->
  <generate>
<artifact output_ref="json_table"/>
  </generate>

  <!-- Behavior on failure -->
  <fallback>
    <policy>If either input XML is invalid, generate JSON listing which elements are missing in an errors array (for example, missing mapped_category/@id or empty activities), with no header or footer (e.g., {"errors":[...]}).</policy>
  </fallback>
</POML>
\end{lstlisting}

\end{document}